\documentclass[10pt,twocolumn,letterpaper]{article}

\usepackage{cvpr}
\usepackage{graphicx}
\usepackage{caption}
\usepackage{float}
\usepackage{hyperref}
\usepackage{multirow} 

\title{6D Pose Estimation via Keypoint Heatmap Regression with RGB-D Residual Neural Networks
}

\author{
Ismail Aljosevic\\
\textit{Politecnico di Torino}\\
\texttt{\normalsize ismail.aljosevic@studenti.polito.it}
\and
Amir Masoud Almasi\\
\textit{Politecnico di Torino}\\
\texttt{\normalsize amirmasoud.almasi@studenti.polito.it}
\and
Ana Parovic\\
\textit{Politecnico di Torino}\\
\texttt{\normalsize ana.parovic@studenti.polito.it}
\and
Ashkan Shafiei\\
\textit{Politecnico di Torino}\\
\texttt{\normalsize ashkan.shafiei@studenti.polito.it}
}

\begin{document}
\maketitle

\begin{abstract}
In this paper, we propose a modular framework for 6D pose estimation based on keypoint heatmap regression. Our approach combines YOLOv10m for object detection with a ResNet18-based network that predicts 2D heatmaps from RGB images. Keypoints extracted from these heatmaps are used to estimate the 6D object pose via the PnP RANSAC algorithm. We compare different keypoint selection strategies to assess their impact on pose accuracy. Additionally, we extend the baseline by incorporating depth data using a cross-fusion architecture, which enables interaction between RGB and depth features at multiple stages. We further explore general training improvements, such as
experimenting with activation functions and learning rate
scheduling strategies to improve model performance. Our best RGB-only model achieved a mean ADD-based accuracy of 84.50\%, while the RGB-D fusion model reached 92.41\% on the LINEMOD dataset. The code is available at \url{https://github.com/ameermasood/HeatNet}.
\end{abstract}
\section{Introduction}

6D pose estimation refers to the task of determining the position and orientation of an object in 3D space, typically using RGB or RGB-D image data. It is a key task in many areas of computer vision, especially robotics, augmented reality, and autonomous systems, where understanding how an object is positioned is crucial for interaction and navigation.

The problem of 6D object pose estimation has traditionally been approached through template-based methods, where the goal is to find the most similar template by matching a set of pre-rendered 2D images of a 3D object to the target scene. While effective in some cases, this approach requires distinguishable surface patterns or textures to achieve reliable template matching, which limits its applicability to texture-less or reflective objects.

In recent years, deep learning has enabled the development of various end-to-end approaches for 6D pose estimation. Among these, regression-based and feature-based methods have demonstrated notable improvements over traditional techniques. Nevertheless, the problem remains highly challenging, particularly in cluttered scenes, under occlusion, or when objects share similar visual appearances.

In this work, we present a feature-based deep learning framework for 6D pose estimation that detects 2D keypoints via heatmap regression and computes the final pose using the \textit{Perspective-n-Point} (PnP) algorithm. We first establish a baseline using RGB images only and then extend the approach by incorporating depth information. Additionally, we evaluate the impact of different 3D keypoint selection strategies (\textit{Farthest Point Sampling} (FPS) and \textit{Curvature Point Sampling} (CPS)), activation functions (ReLU, SiLU, and Mish), and learning-rate schedulers (OneCycleLR and PolynomialLR) on overall model performance.

\begin{figure*}[t]
    \centering
    \includegraphics[width=\textwidth]{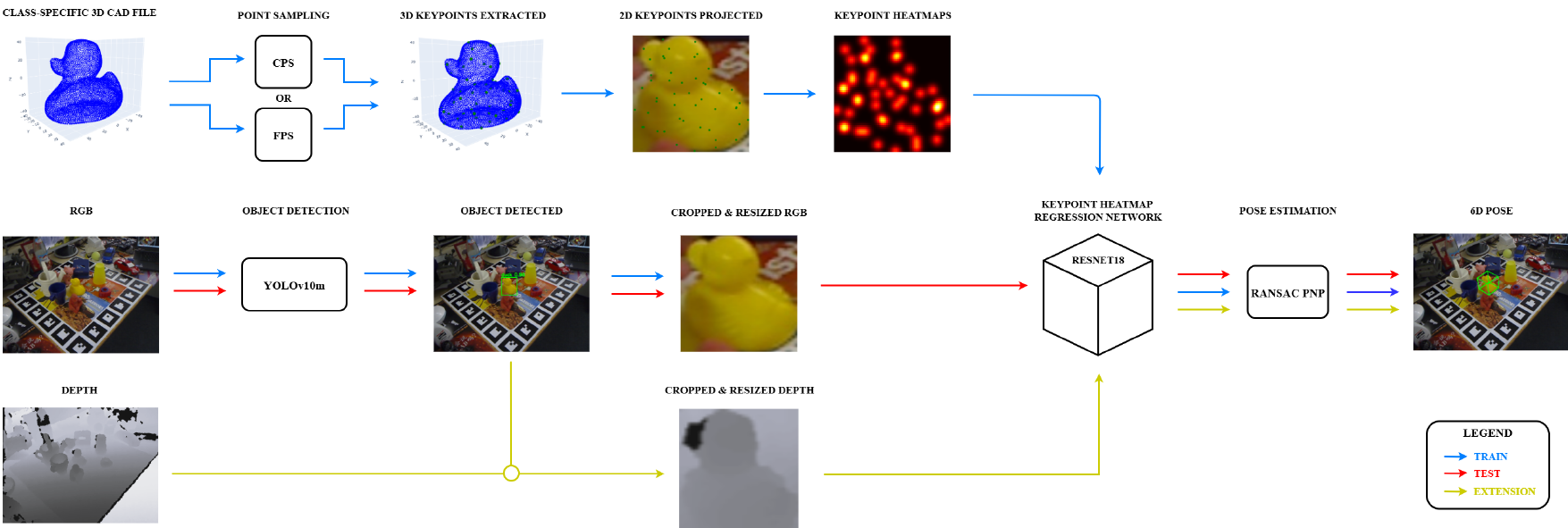}
    \caption{Pipeline overview with a clear separation of the training process (blue), test flow (red), and depth extension (yellow).}
    \label{fig:overall-pipeline}
\end{figure*}
\section{Related Work}

This section reviews several methods that have recently driven progress in 6D pose estimation, with a particular focus on feature-based approaches for keypoint detection and object pose estimation.

Zhao et al.~\cite{zhao2018estimating} proposed a two-stage RGB-only pipeline that uses YOLOv3 for object detection and a ResNet-101-based network for heatmap-based keypoint localization, followed by PnP for final pose estimation. The method achieves strong results without requiring refinement, but it shows limitations for symmetric or texture-less objects.

PVNet~\cite{peng2019pvnet} introduced a voting-based approach in which the network predicts unit vectors from each pixel toward predefined 3D keypoints projected onto the image plane. Keypoints are selected using FPS, and semantic segmentation is used to isolate object regions. The method is robust to occlusions and to symmetric and texture-less objects, although the voting and segmentation stages increase computational complexity.

Aing and Lie~\cite{aing2021detecting} introduced CPS, which selects keypoints from high-curvature regions to improve localization by focusing on distinctive geometric areas. Combined with a voting mechanism similar to that of PVNet, this approach improves robustness under occlusion. However, it struggles with symmetric objects, where keypoints from different views may appear indistinguishable.

Finally, Sun et al.~\cite{sun2021crossfunet} proposed CrossFuNet for 3D human pose estimation, which integrates depth information through a cross-fusion mechanism. Separate subnetworks for RGB and depth data interact at multiple levels, improving robustness to self-occlusion and depth ambiguity, but increasing architectural complexity and dependence on depth sensors.
\section{Methodology}

This section describes the pipeline used for 6D pose estimation in more detail, including the dataset setup, configuration details, baseline assessment, and the proposed depth extension.

\subsection{Architecture Overview}
The overall pipeline is illustrated in Fig.~\ref{fig:overall-pipeline}, showing a clear separation between the training process, testing flow, and depth extension.

The pipeline begins with object detection on the RGB image using YOLOv10m to locate the region containing the target object. This region is then cropped from the original image, allowing the following stages to focus exclusively on the target object.

The next stage focuses on detecting object keypoints in the cropped image, which is achieved indirectly through heatmap estimation. For this purpose, separate data preparation is required. Initially, it is necessary to determine which keypoints should be detected in the images for each object. These keypoints are selected from the provided CAD models using two sampling strategies: FPS and CPS. The selected keypoints are then projected onto the image plane to define ground truth locations. These ground truth locations are used to generate heatmaps, which serve as training targets for the keypoint heatmap regression network.

After the keypoint heatmap regression step, a set of heatmaps is generated for the input image. The keypoints are obtained by locating the pixel with the maximum activation in each heatmap. These predicted keypoints, along with corresponding 3D points initially sampled from the CAD model, serve as input to the PnP algorithm to estimate the object's rotation and translation.

As part of the extended architecture, depth information is integrated into the keypoint prediction pipeline. The region of interest detected by YOLO is cropped from both RGB and the corresponding depth images, which are then processed separately by a dual-stream ResNet-based model.

\subsection{Dataset}

The dataset used in our project is a subset of the LINEMOD dataset, a popular benchmark for 6D object pose estimation~\cite{hinterstoisser2012model}. LINEMOD includes 15 texture-less object classes captured in real-world environments under varying poses, occlusions, and lighting conditions. Each object instance is provided with a 3D mesh model, RGB images, depth maps, masks, and corresponding ground-truth labels for bounding boxes and poses.

We utilized 15,800 RGB images from 13 object classes, since classes 03 and 07 were excluded due to inconsistencies in annotations. The dataset was split into 90\% for training (14{,}220 images) and 10\% for testing (1{,}580 images).

\subsection{Object Detection}

For the object detection phase, we used the YOLOv10m model to locate and classify the object present in each RGB image. YOLOv10m is a recent version of the YOLO (You Only Look Once) family of object detectors~\cite{yolov10}.

YOLO is a single-stage object detector, meaning it performs both object localization and classification in a single pass through the network. The model divides the input image into a grid, and each grid cell is responsible for predicting bounding box coordinates, the class of the object, and a confidence score. This confidence score reflects the model’s certainty that an object of the predicted class is present within the proposed region.

The model is trained by minimizing a weighted multi-term loss:
\[
\mathcal{L} = \lambda_{\text{loc}} \mathcal{L}_{\text{loc}} + \lambda_{\text{conf}} \mathcal{L}_{\text{conf}} + \lambda_{\text{cls}} \mathcal{L}_{\text{cls}}
\]

where \( \lambda_{\text{loc}} \), \( \lambda_{\text{conf}} \), and \( \lambda_{\text{cls}} \) are hyperparameters controlling the contributions of the localization, confidence, and classification terms, respectively.

The detection performance is evaluated using the \textit{mean Average Precision} (mAP), defined as:

\[
\text{mAP} = \frac{1}{C} \sum_{c=1}^{C} \text{AP}_c
\]

where \( C \) denotes the number of object classes and \( \text{AP}_c \) is the area under the precision-recall curve for class \( c \).

After training, YOLO predictions are used to crop object regions from the RGB images, which are subsequently resized to $256 \times 256$ pixels to ensure consistent input dimensions for the next stage. Furthermore, we used YOLO detections to generate cropped training images for the following stage of the pipeline. This approach allows the subsequent model to be trained directly on data that includes the errors introduced during the object detection phase, thus improving robustness.

\subsection{3D Keypoints Sampling}
Before the keypoint detection phase, the set of keypoints that must be detected on each object was first determined. To select these keypoints from the provided CAD models, we used two 3D sampling strategies: \textit{Farthest Point Sampling} (FPS) and \textit{Curvature Point Sampling} (CPS).

\textbf{Farthest Point Sampling (FPS):}
FPS selects a set of points that are maximally spread across the object surface. Starting from a random point, it iteratively adds the point farthest from the current set, ensuring uniform spatial coverage.

\textbf{Curvature Point Sampling (CPS):}
CPS selects points from regions of the object surface with the highest curvature, corresponding to geometrically complex or detailed areas.
First, the curvature at each point on the 3D model is estimated based on how much the surface bends around it. This is achieved by computing the eigenvalues of the covariance matrix of its \(k\)-nearest neighbors. The curvature for point \(p_i\) is computed as:

\[
\kappa(p_i) = \frac{\lambda_0}{\lambda_0 + \lambda_1 + \lambda_2 + \varepsilon}
\]

where \( \lambda_0, \lambda_1, \lambda_2 \) are the eigenvalues sorted in ascending order, and \(\varepsilon\) is a small constant to avoid division by zero.

To encourage diversity and avoid the selection of keypoints near the object center, a final score is assigned to each point based on its curvature and its distance from the object center \( \bar{p} \):

\[
\text{score}(p_i) = \kappa(p_i) \cdot \|p_i - \bar{p}\|_2.
\]
Finally, the points with the highest scores are selected.

For both FPS and CPS, we selected 50 keypoints per object to balance geometric detail and efficiency. The selected 3D keypoints are projected onto the image plane, relative to the cropped images, to define ground truth locations. These locations are used to generate heatmaps for training the keypoint regression network.

\subsection{Keypoint Heatmap Regression}

Although direct coordinate regression may seem more intuitive, heatmap-based methods offer a more structured and informative approach. Instead of regressing coordinates directly, the model learns to generate a 2D probability distribution—a heatmap—for each keypoint, where each value represents the likelihood of the keypoint being present at a particular pixel location. This spatial representation helps the model focus on the local neighborhood around a keypoint, rather than treating its position as an isolated coordinate.
Furthermore, learning to predict dense heatmaps supplies the network with rich spatial gradients during training, resulting in more stable optimization and faster convergence~\cite{zhao2018estimating}.

\begin{figure*}[t]
    \centering
    \includegraphics[width=\textwidth]{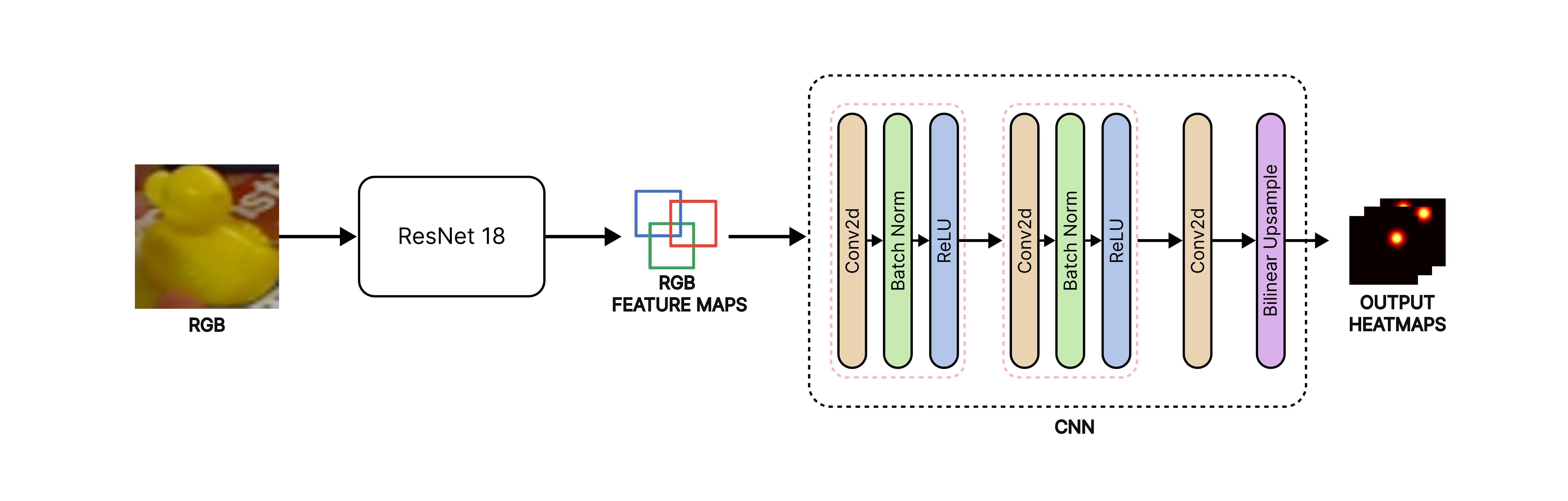}
    \caption{
    Overview of the heatmap regression network.
    }
    \label{fig:heatmap_network}
\end{figure*}

Motivated by these advantages, we adopt a convolutional neural network to predict heatmaps from cropped RGB images. The architecture, illustrated in Fig.~\ref{fig:heatmap_network}, is based on a pre-trained ResNet-18 backbone for feature extraction, followed by a lightweight decoder that outputs one heatmap per keypoint. ResNet-18 is well-suited for this task, as it extracts rich features while keeping computational cost low, making it effective for dense prediction tasks like keypoint heatmap regression.

As training data for the network, we use ground truth heatmaps generated from the previously obtained 2D keypoints. Each heatmap is created by applying a Gaussian kernel centered at the corresponding keypoint location. Formally, for each keypoint with 2D coordinates \((x, y)\), a corresponding heatmap \(H(x', y')\) is generated as:

\[
H(x', y') = \exp\left(-\frac{(x' - x)^2 + (y' - y)^2}{2\sigma^2}\right),
\]

where we set \(\sigma = 2.0\) to control the spatial spread of the Gaussian.

Unlike the input RGB images obtained from the YOLO output, which have a resolution of \(256 \times 256\), the predicted heatmaps are generated at a lower resolution of \(64 \times 64\). This downsampling is a standard practice in keypoint detection pipelines, as it reflects the spatial resolution of the final feature maps produced by the backbone network.

Since we previously selected 50 keypoints per object, the output of the network is a tensor of shape \([50, 64, 64]\), where each channel corresponds to the likelihood map of a specific keypoint.

\subsection{Pose Estimation}

After obtaining the heatmaps from the keypoint heatmap network, the 2D keypoints are extracted by identifying the position of the pixel with the highest activation in each heatmap. Since the heatmaps are at a lower resolution, the predicted keypoint coordinates are first scaled to match the cropped image size and then mapped back to the coordinates of the original image.

The resulting 6D pose consists of a rotation matrix and a translation vector, which together describe the object's pose in the camera coordinate system. To estimate this pose from the 2D--3D correspondences, the predicted 2D keypoints are paired with their corresponding 3D points, previously sampled from the CAD model using FPS and CPS. We apply the \textit{Perspective-n-Point} (PnP) algorithm in combination with RANdom SAmple Consensus (RANSAC). PnP estimates the object’s rotation and translation by aligning the 3D points with their 2D projections onto the image plane. To improve robustness against noisy predictions and outliers, RANSAC iteratively generates pose hypotheses and selects the solution with the highest number of inliers. In this context, inliers are keypoints whose projected positions fall within an acceptable error margin of their detected 2D locations, based on a predefined reprojection error threshold.

\begin{figure*}[t]
    \centering
    \includegraphics[width=\textwidth]{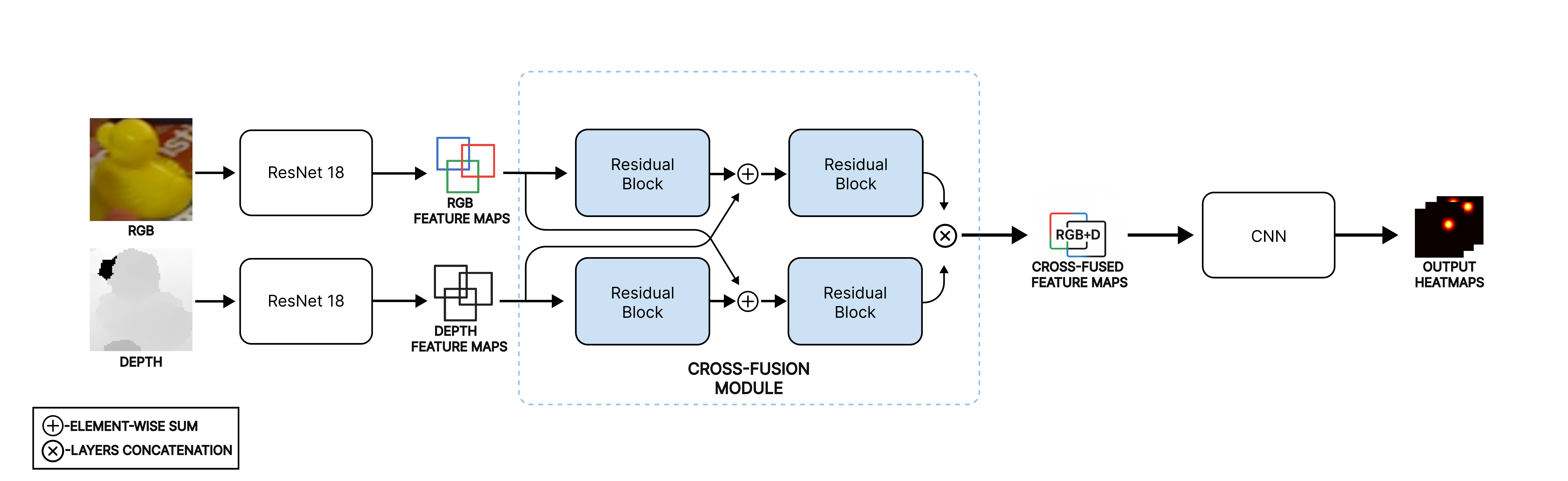}
    \caption{
        Overview of the proposed extended architecture, which combines RGB and depth features to improve pose estimation.
        }
    \label{fig:cross-fusion}
\end{figure*}

\subsection{Depth Extension}

As an extension of the baseline approach, we incorporate a cross-fusion strategy into the architecture. The extended design introduces depth information alongside RGB input via a dual-stream setup. RGB and depth images are processed separately using their own ResNet-18 networks. This allows each network to capture features specific to each modality. These separate features are then combined through a fusion module, creating a unified representation that leverages the strengths of both RGB and depth data. The extended architecture, with emphasis on the cross-fusion module, is illustrated in Fig.~\ref{fig:cross-fusion}.

The cross-fusion mechanism is implemented using residual blocks. Feature maps from the two branches are first passed through separate residual blocks. The output of the RGB branch is then element-wise added to the original depth feature maps, while the output of the depth branch is added to the original RGB feature maps. This bidirectional interaction allows both modalities to refine their representations using information from the other~\cite{sun2021crossfunet}.

The resulting feature maps are then processed by a second set of residual blocks in each branch. Finally, the outputs from both branches are concatenated to form the cross-fused feature maps, which are forwarded to a CNN decoder identical to the one used in the baseline architecture.

Unlike early fusion methods that treat depth as an additional input channel, the proposed cross-fusion architecture enables multi-level, bidirectional exchange between RGB and depth streams. This allows the network to combine texture and color from RGB with geometric information from depth, resulting in more informative features. Moreover, the use of residual connections within the fusion module enhances robustness and spatial precision, leading to more accurate keypoint predictions.

\subsection{Training Configuration}

The initial training configuration, shared by both the baseline and the extended model, is summarized in Table~\ref{tab:training_config}. Both models are trained using the Adam optimizer with a fixed learning rate. ReLU is employed as the activation function across all layers. As the loss function, we use a \textit{Focal Heatmap Loss}.

\begin{table}[h]
\centering
\begin{tabular}{l|l}
\hline
\textbf{Parameter} & \textbf{Value} \\
\hline
Optimizer & Adam \\
Learning rate & \(1 \times 10^{-4}\) \\
Batch size & 16 \\
Learning rate scheduler & None \\
Number of epochs & 30 \\
Activation function & ReLU \\
Loss function & Focal Heatmap Loss \\
\hline
\end{tabular}
\caption{Initial training configuration for both baseline and extended architectures}
\label{tab:training_config}
\end{table}

Focal Heatmap Loss is a variant of focal loss designed to address the severe class imbalance in keypoint heatmap regression, where keypoints occupy only a small region of the heatmap while most pixels correspond to background \cite{zhou2019objects}. Standard losses such as MSE are dominated by background pixels, reducing the sensitivity to keypoint regions. To overcome this, Focal Heatmap Loss down-weights easy negatives and focuses the learning on hard, informative pixels near keypoints:

\[
\mathcal{L}_{\text{focal}}(p, y) =
\begin{cases}
(1 - p)^{\gamma} \cdot \log(p) & \text{if } y = 1 \\
(1 - y)^{\beta} \cdot p^{\gamma} \cdot \log(1 - p) & \text{if } y < 1
\end{cases}
\]

where $p$ is the predicted heatmap value, $y$ is the ground truth, $\gamma$ controls the focus on hard examples, and $\beta$ adjusts the background penalty.

The total loss is computed by averaging the pixel-wise loss across all keypoints, spatial locations, and batch samples:

\[
\mathcal{L}_{\text{total}} = \frac{1}{N} \sum \mathcal{L}_{\text{focal}}(p, y)
\]

where $N$ denotes the total number of pixels across the batch.

\subsection{Baseline Assessment}

As a starting point, we adopt a baseline model, focusing on the impact of FPS and CPS selection strategies. All models are evaluated using the Average Distance of Model Points (ADD) metric, which measures the average distance between model points transformed by the predicted pose $(\hat{R}, \hat{\mathbf{t}})$ and the ground truth pose $(R, \mathbf{t})$:

\[
\text{ADD} = \frac{1}{|\mathcal{M}|} \sum_{\mathbf{x} \in \mathcal{M}} \left\| R \mathbf{x} + \mathbf{t} - \hat{R} \mathbf{x} - \hat{\mathbf{t}} \right\|.
\]

For symmetric objects, we use the ADD-S variant, which replaces point-wise correspondence with closest-point matching:

\[
\text{ADD-S} = \frac{1}{|\mathcal{M}|} \sum_{\mathbf{x}_1 \in \mathcal{M}} \min_{\mathbf{x}_2 \in \mathcal{M}} \left\| R \mathbf{x}_1 + \mathbf{t} - \hat{R} \mathbf{x}_2 - \hat{\mathbf{t}} \right\|.
\]

Symmetric objects are defined as those whose geometry exhibits invariance under certain rotations or reflections, resulting in pose ambiguities when evaluated using standard distance metrics.

\begin{figure}[h]
    \centering
    \begin{minipage}{0.48\linewidth}
        \centering
        \includegraphics[width=\linewidth]{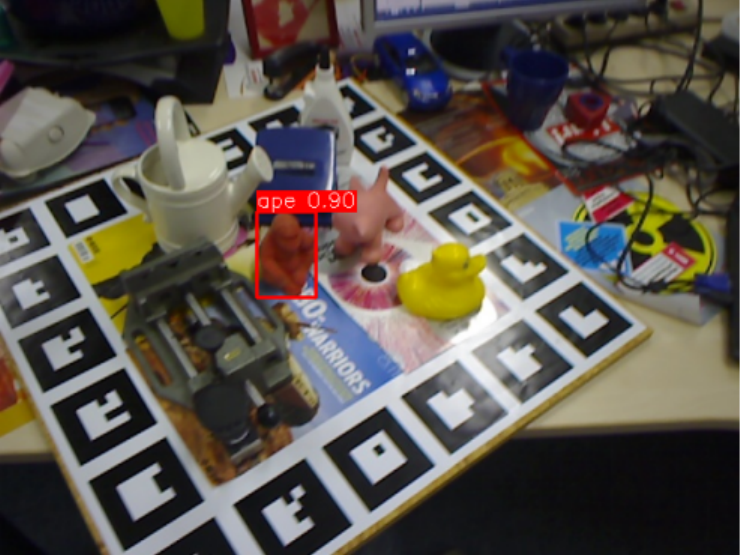}
        \vspace{-0.3cm}
        \subcaption{YOLOv10m Object Detection}
    \end{minipage}
    \hfill
    \begin{minipage}{0.48\linewidth}
        \centering
        \includegraphics[width=\linewidth]{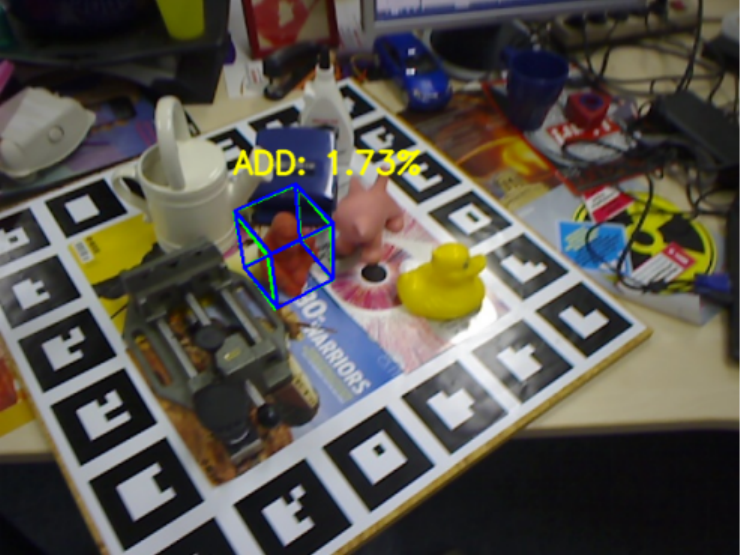}
        \vspace{-0.3cm}
        \subcaption{6D pose estimation}
    \end{minipage}

    \caption{Qualitative results of YOLOv10m object detection (left) and the final 6D pose estimation of the detected object (right).}
    \label{fig:yolo_6dpose}
\end{figure}

A pose prediction is considered correct if the ADD (or ADD-S) error is below 10\% of the object’s diameter. The results obtained for the baseline models using both keypoint sampling strategies are shown in Table~\ref{tab:baseline_results}, where symmetric objects are marked with an asterisk (*). The table also includes YOLO detection results, reported in terms of mAP. In addition, qualitative examples of object detection and 6D pose estimation are presented in Fig.~\ref{fig:yolo_6dpose}.

\begin{table}[h]
\centering
\small
\begin{tabular}{l|c|cc}
\hline
\multirow{2}{*}{\textbf{\shortstack{Object \\ Title}}} & \multicolumn{1}{c|}{\textbf{YOLOv10m}} & \multicolumn{2}{c}{\textbf{6D pose accuracy}} \\
 & \textbf{mAP@50:95} & \textbf{CPS} & \textbf{FPS} \\
\hline
ape & 0.961 & 52.61\% & 61.14\% \\
bench vise & 0.981 & 93.18\% & 93.64\% \\
camera & 0.976 & 87.32\% & 89.27\% \\
can & 0.989 & 91.96\% & 89.73\% \\
cat & 0.980 & 78.38\% & 81.98\% \\
driller & 0.980 & 95.45\% & 97.47\% \\
duck & 0.967 & 57.78\% & 66.67\% \\
eggbox* & 0.977 & 81.58\% & 87.72\% \\
glue* & 0.945 & 86.49\% & 91.44\% \\
hole puncher & 0.970 & 85.58\% & 87.91\% \\
iron & 0.984 & 91.71\% & 93.09\% \\
lamp & 0.978 & 96.43\% & 98.21\% \\
phone & 0.979 & 84.91\% & 88.79\% \\
\hline
\textbf{Mean} & \textbf{0.974} & \textbf{83.34\%} & \textbf{86.70\%} \\
\hline
\end{tabular}
\caption{Validation performance for YOLOv10m object detection (mAP@50:95) and 6D pose estimation using CPS and FPS keypoint sampling strategies.}
\label{tab:baseline_results}
\end{table}

\begin{figure}[h]
    \centering
    \includegraphics[width=0.7\linewidth]{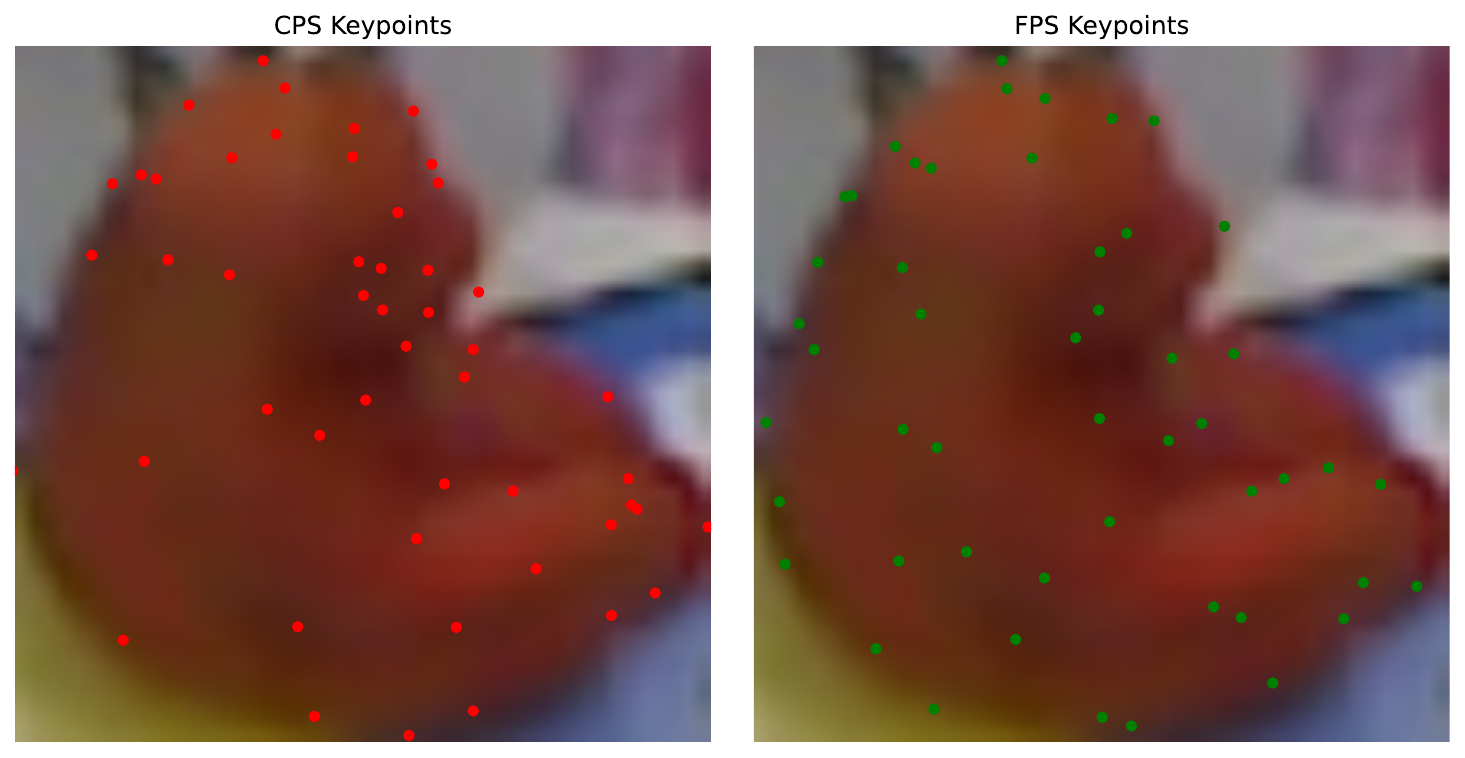}
\caption{Projected keypoints using CPS (left) and FPS (right). CPS focuses on specific surface regions and ignores flatter areas such as the back of the \textit{ape} object, while FPS provides a more uniform coverage of the entire surface.}
    \label{fig:fps_vs_cps}
\end{figure}

The YOLO detection module achieved consistently high performance in our experiments, resulting in minimal error contribution to the overall 6D pose estimation. Consequently, the reported 6D pose accuracy primarily reflects the precision of the keypoint heatmap regression model. The FPS and CPS sampling strategies produced similar results for most object classes. However, the most notable improvements with FPS are observed for challenging objects in the LINEMOD dataset, including symmetric objects such as glue and eggbox, as well as low-texture or small objects such as ape and duck. CPS selects keypoints from the most curved areas of the object surface, which means it tends to ignore flatter regions that may still be informative for pose estimation. In contrast, FPS distributes keypoints more uniformly across the entire surface, ensuring that both detailed and flat regions are represented, which leads to better robustness in these difficult cases. An illustrative example of these differences is provided in Fig.~\ref{fig:fps_vs_cps}.

\section{Experiments and Results}

In this section, we analyze the effect of integrating depth information into the baseline model, as well as additional experiments involving general training improvements. Specifically, we evaluate the impact of using two different activation functions, \textit{SiLU} and \textit{Mish}, and examine the effect of introducing learning rate schedulers (OneCycleLR and PolynomialLR). Since FPS point selection yielded more consistent results, all subsequent experiments are conducted using this strategy.

The results of the extended model, trained using the initial configuration from Table~\ref{tab:training_config}, are presented in Table~\ref{tab:val_pose_full_width} under the ReLU column. As expected, incorporating depth information improved overall performance, with an average accuracy of 88.19\%. Most object classes either maintained or improved their performance with the addition of depth. In contrast, a slight performance drop was observed for a few classes, including \textit{duck}, \textit{can}, and \textit{camera}. Although these differences are relatively small, they were unexpected and may be due to noise or misalignment in the depth data.

\begin{table*}[h]
\centering
\small
\setlength{\tabcolsep}{8pt}
\renewcommand{\arraystretch}{1}
\resizebox{\textwidth}{!}{
\begin{tabular}{l|ccc|ccc|ccc}
\hline
\multirow{2}{*}{\textbf{\shortstack{Object \\ Title}}} & \multicolumn{3}{c|}{\textbf{ConstantLR}} & \multicolumn{3}{c|}{\textbf{OneCycleLR}} & \multicolumn{3}{c}{\textbf{PolynomialLR}} \\
 & \textbf{ReLU} & \textbf{SiLU} & \textbf{Mish} & \textbf{ReLU} & \textbf{SiLU} & \textbf{Mish} & \textbf{ReLU} & \textbf{SiLU} & \textbf{Mish} \\
\hline
ape           & 67.30\% & 60.19\% & 65.88\% & 72.04\% & 72.51\% & 72.04\% & 67.77\% & 64.93\% & 68.72\% \\
bench vise    & 96.36\% & 91.36\% & 94.55\% & 99.09\% & 95.45\% & 96.82\% & 95.45\% & 96.82\% & 95.45\% \\
camera        & 92.20\% & 87.32\% & 88.78\% & 95.12\% & 94.15\% & 95.61\% & 89.76\% & 91.22\% & 93.66\% \\
can           & 91.52\% & 83.93\% & 90.63\% & 93.30\% & 95.09\% & 94.20\% & 89.73\% & 88.39\% & 87.95\% \\
cat           & 86.94\% & 74.32\% & 85.59\% & 86.94\% & 91.44\% & 90.99\% & 86.94\% & 83.78\% & 85.14\% \\
driller       & 96.97\% & 92.42\% & 96.46\% & 99.49\% & 97.47\% & 98.48\% & 96.97\% & 98.48\% & 97.47\% \\
duck          & 65.33\% & 62.67\% & 66.22\% & 66.22\% & 73.33\% & 79.11\% & 67.56\% & 65.78\% & 65.33\% \\
eggbox*       & 82.89\% & 87.72\% & 90.35\% & 89.47\% & 92.54\% & 91.67\% & 87.72\% & 82.89\% & 85.09\% \\
glue*         & 92.34\% & 89.64\% & 95.50\% & 89.64\% & 95.50\% & 95.95\% & 92.34\% & 90.54\% & 91.44\% \\
hole puncher  & 87.91\% & 78.60\% & 88.37\% & 78.60\% & 88.37\% & 90.23\% & 83.72\% & 87.44\% & 83.26\% \\
iron          & 95.39\% & 90.32\% & 95.39\% & 95.39\% & 95.85\% & 96.77\% & 92.17\% & 90.32\% & 92.17\% \\
lamp          & 98.66\% & 95.54\% & 99.11\% & 99.11\% & 99.11\% & 100.00\% & 96.43\% & 98.66\% & 97.32\% \\
phone         & 92.67\% & 83.19\% & 91.38\% & 92.24\% & 92.24\% & 92.67\% & 89.66\% & 90.52\% & 87.93\% \\
\hline
\textbf{Mean} & \textbf{88.19\%} & \textbf{82.86\%} & \textbf{88.32\%} & \textbf{90.14\%} & \textbf{91.08\%} & \textbf{91.92\%} & \textbf{87.40\%} & \textbf{86.91\%} & \textbf{86.99\%} \\
\hline
\end{tabular}
}
\caption{6D pose validation accuracy (\%) of the extended model with different activations and learning rate schedulers.}
\label{tab:val_pose_full_width}
\end{table*}

\subsection{General Training Improvements}
\subsubsection{Activation Functions}

The first modification focuses on evaluating the impact of different activation functions on model performance. In particular, we experimented with \textit{SiLU} (Sigmoid Linear Unit) and \textit{Mish}, both of which are smooth, non-monotonic functions designed to improve gradient flow and feature representation.

\textbf{SiLU} is defined as:
\[
\text{SiLU}(x) = x \cdot \sigma(x)
\]
where \(\sigma(x)\) denotes the sigmoid function. SiLU provides a smooth and bounded activation that helps avoid dead neurons and supports more stable optimization~\cite{misra2020mishselfregularizednonmonotonic}.

\textbf{Mish} is defined as:
\[
\text{Mish}(x) = x \cdot \tanh(\ln(1 + e^x)).
\]
Mish is a self-regularized non-monotonic activation function that preserves small negative values. It has been shown to outperform traditional activations such as ReLU and even SiLU in various vision tasks by enabling better gradient flow and smoother output transitions~\cite{misra2020mishselfregularizednonmonotonic}.

The results obtained by applying the \textit{SiLU} and \textit{Mish} activation functions are presented in Table~\ref{tab:val_pose_full_width}. Compared to the configuration with ReLU, the Mish activation achieved similar results, with slightly better performance on symmetric objects. However, the use of SiLU led to a significant accuracy drop across most objects, which may be attributed to its higher sensitivity to parameter selection, including learning rate, batch size, and other training hyperparameters.

\subsubsection{Learning Rate Scheduling}

The second modification explores how learning rate scheduling affects model performance. Schedulers adjust the learning rate during training to help the model converge faster and more effectively. We experimented with two types of schedulers: OneCycleLR and PolynomialLR.

The \textbf{OneCycleLR} scheduler belongs to the group of cycle-based schedulers, where the learning rate increases and then decreases within a single training cycle, following predefined functions such as linear or cosine decay. This type of schedule helps the model escape poor local minima and accelerates convergence.

The \textbf{PolynomialLR} scheduler belongs to the group of decay-based schedulers, where the learning rate is gradually reduced over time. It helps stabilize training and improve final accuracy.

We applied two learning rate scheduler configurations—OneCycleLR with cosine annealing decay and PolynomialLR with quadratic decay—across different activation functions. The results are presented in Table~\ref{tab:val_pose_full_width}.

The configuration with Mish activation and the OneCycleLR scheduler achieved the best overall performance, confirming the positive impact of dynamic learning rate adjustment on model generalization. In contrast, when using PolynomialLR, the results with Mish showed a noticeable drop in accuracy. This can be attributed to the more aggressive nature of the quadratic decay, which may have reduced the learning rate too quickly, limiting the model's ability to optimize in the later stages.

Interestingly, the use of SiLU in combination with both schedulers resulted in significant improvements compared to the baseline, further confirming the known sensitivity of this activation function to parameter selection, particularly the learning rate.

The results with ReLU followed a similar trend to Mish, where the combination with OneCycleLR provided better performance compared to PolynomialLR. This further suggests that aggressive learning rate decay, as seen with PolynomialLR, can obstruct the model's ability to fully converge, regardless of the chosen activation function.

The test results for the best baseline and extended models are shown in Table~\ref{tab:test_results}, confirming the trends observed during validation. Among all configurations, the combination of ReLU activation and the OneCycleLR scheduler achieved the best performance. Although the mean test accuracy for the extended model slightly exceeds the validation performance, which may appear atypical, this can be attributed to random variations in the complexity of test samples.

\begin{table}[h]
\centering
\begin{tabular}{l|c|c}
\hline
\textbf{\shortstack{Object \\ Title}} & \textbf{\shortstack{Baseline \\ (ReLU)}}
 & \textbf{\shortstack{Extended \\ (ReLU + OneCycleLR)}}
 \\
\hline
ape           & 52.42\% & 73.39\% \\
bench vise    & 92.56\% & 97.52\% \\
camera        & 90.83\% & 92.50\% \\
can           & 90.00\% & 93.33\% \\
cat           & 77.12\% & 93.22\% \\
driller       & 96.64\% & 98.32\% \\
duck          & 61.60\% & 80.00\% \\
eggbox*       & 87.20\% & 96.00\% \\
glue*         & 86.89\% & 95.90\% \\
hole puncher  & 83.87\% & 93.55\% \\
iron          & 93.91\% & 95.65\% \\
lamp          & 95.90\% & 96.72\% \\
phone         & 89.52\% & 95.16\% \\
\hline
\textbf{Mean} & \textbf{84.50\%} & \textbf{92.41\%} \\
\hline
\end{tabular}
\caption{Test results of the best baseline and extended models.}
\label{tab:test_results}
\end{table}
\section{Discussion and Findings}
In this work, we analyzed the impact of several factors relevant to 6D pose estimation, including keypoint sampling strategies, depth integration, and general training improvements.

From the baseline experiments, we observed that the keypoint sampling strategies yielded similar results across most object classes. However, FPS showed a clear advantage on several challenging objects, which justified its use in all subsequent experiments and helped explain why it is more commonly used in related work.

We further explored depth integration using a dual-stream ResNet-18 architecture with cross-fusion. This extension improved performance across most object classes; however, it did not fully resolve the challenges observed on difficult LINEMOD objects such as \textit{ape} and \textit{duck}.

Finally, experiments with general training improvements confirmed the importance of selecting an appropriate configuration, as even changes in activation functions or learning rate schedules affected pose estimation performance. Notably, these improvements were most visible on the more challenging LINEMOD objects, contributing to higher robustness, although certain limitations remain.

Due to limited computational resources, some analyses were not conducted but remain important for future work. In particular, we used a large number of keypoints to maximize detection performance, and further study is needed to quantify how keypoint count affects results. Heatmap-based methods typically use more keypoints than pixel-wise voting, improving robustness but potentially introducing noise; although RANSAC filters unreliable correspondences within PnP, tuning the number of selected keypoints could further improve accuracy and stability. In addition, advanced data augmentation and deeper network architectures could further improve performance, especially for challenging object classes where the current models still show lower accuracy and reduced robustness.

In conclusion, this work integrates different aspects from multiple existing approaches into a unified 6D pose estimation pipeline. The proposed approach provides a solid foundation for further experimentation and future improvements.

\section*{Acknowledgments}
This work was carried out in the context of the Machine Learning and Deep Learning course at Politecnico di Torino, Italy. We thank Barbara Caputo, Raffaello Camoriano, Stephany Ortuno Chanelo, and Paolo Rabino for their guidance and support.

{\small
\bibliographystyle{ieeenat_fullname}
\bibliography{main}
}

\end{document}